\documentclass{article}
\usepackage[preprint]{neurips_2020}
\pdfoutput=1
\usepackage{algpseudocode,algorithm,algorithmicx}
\usepackage{graphicx}
\usepackage[utf8]{inputenc} % allow utf-8 input
\usepackage[T1]{fontenc}    % use 8-bit T1 fonts
\usepackage{hyperref}       % hyperlinks
\usepackage{url}            % simple URL typesetting
\usepackage{booktabs}       % professional-quality tables
\usepackage{amsfonts}       % blackboard math symbols
\usepackage{nicefrac}       % compact symbols for 1/2, etc.
\usepackage{microtype}      % microtypography

\usepackage{soul}
\usepackage{booktabs} % for professional tables
\usepackage{times}
\usepackage{epsfig}
\usepackage{dsfont}
\usepackage{amsmath}
\usepackage{amssymb}
\usepackage{tabularx}
\usepackage{makecell}
\usepackage{bm}
\usepackage{multirow} 
\usepackage{rotating}
\usepackage{wrapfig}
\usepackage{subfigure}
\usepackage{amsthm}
\usepackage{thmtools}
\usepackage{thm-restate}
\usepackage{flexisym}
\usepackage{colortbl}
\usepackage{xcolor}
\usepackage{appendix}
\usepackage{caption}

\newcolumntype{C}[1]{>{\centering\arraybackslash}p{#1}}

\definecolor{adv_red}{RGB}{225,100,100}
\definecolor{adv_blue}{RGB}{60,120,215}

\graphicspath{{./images/}}

\usepackage[symbol]{footmisc}
\renewcommand{\thefootnote}{\fnsymbol{footnote}}

\DeclareMathOperator*{\argmin}{arg\,min}

\title{Network Architecture Search for Domain Adaptation}

\author{%
  Yichen Li$^{*}$ \\
  Stanford University \\
  Stanford, CA \\
  \texttt{liyichen@stanford.edu} \\
   \And
   Xingchao Peng$^{*}$ \\
   Boston University \\
   Boston, MA \\
   \texttt{xpeng@bu.edu} \\
}

\begin{document}

\maketitle

\renewcommand{\thefootnote}{\fnsymbol{footnote}}\footnotetext[1]{:indicates equal contributions.}

\begin{abstract}

Deep networks have been used to learn transferable representations for domain adaptation. Existing deep domain adaptation methods systematically employ popular hand-crafted networks designed specifically for image-classification tasks, leading to sub-optimal domain adaptation performance. In this paper, we present Neural Architecture Search for Domain Adaptation (NASDA), a principle framework that leverages differentiable neural architecture search to derive the optimal network architecture for domain adaptation task. NASDA is designed with two novel training strategies: neural architecture search with multi-kernel Maximum Mean Discrepancy to derive the optimal architecture, and adversarial training between a feature generator and a batch of classifiers to consolidate the feature generator. 
We demonstrate experimentally that NASDA leads to state-of-the-art performance on several domain adaptation benchmarks.

\end{abstract}

\section{Introduction}
\label{intro}

Supervised machine learning models ($\Phi$) aim to minimize the empirical test error ($\epsilon (\Phi(\mathbf{x}), \mathbf{y})$) by optimizing $\Phi$ on training data ($\mathbf{x}$) and ground truth labels ($\mathbf{y}$), assuming that the training and testing data are sampled \textit{i.i.d} from the same distribution. While in practical, the training and testing data are typically collected from related domains under different distributions, a phenomenon known as {domain shift} (or domain discrepancy)~\cite{datashift_book2009}. To avoid the cost of annotating each new test data, {Unsupervised Domain Adaptation} (UDA) tackles domain shift by transferring the knowledge learned from a rich-labeled source domain ($P(\mathbf{x}^s, \mathbf{y}^s)$) to the unlabeled target domain ($Q(\mathbf{x}^t)$). Recently unsupervised domain adaptation research has achieved significant progress with techniques like discrepancy alignment~\cite{JAN,ddc,ghifary2014domain,peng2017synthetic,long2015,SunS16a}, adversarial alignment~\cite{xu2019adversarial,cogan,adda,ufdn,DANN,MCD_2018,long2018NIPS_CDAN}, and reconstruction-based alignment~\cite{yi2017dualgan, CycleGAN2017, hoffman2017cycada, kim2017learning}. While such models typically learn feature mapping from one domain ($\Phi(\mathbf{x}^s)$) to another ($\Phi(\mathbf{x}^t)$) or derive a joint representation across domains ($\Phi(\mathbf{x}^s)\otimes\Phi(\mathbf{x}^t)$), the developed models have limited capacities in deriving an optimal neural architecture specific for domain transfer. 

To advance network designs, 
neural architecture search (NAS) automates the net architecture engineering process by reinforcement supervision~\cite{zoph2016neural} or through neuro-evlolution~\cite{Real2019AgingEF}. 
 Conventional NAS models aim to derive neural architecture $\alpha$ along with the network parameters $w$,
 by solving a bilevel optimization problem~\cite{anandalingam1992hierarchical}: $\Phi_{\alpha, w}= \argmin_{\alpha} \mathcal{L}_{val}(w^*(\alpha), \alpha) $
	$\text{s.t. } w^*(\alpha) = \mathrm{argmin}_w  \mathcal{L}_{train}(w, \alpha) $, where $\mathcal{L}_{train}$ and $\mathcal{L}_{val}$ indicate the training and validation loss, respectively.
While recent works demonstrate competitive performance on tasks such as image classification~\cite{zoph2018learning,liu2017hierarchical,liu2018progressive,real2019regularized} and object detection~\cite{zoph2016neural}, designs of existing NAS algorithms typically assume that the training and testing domain are sampled from the same distribution, neglecting the scenario where two data domains or multiple feature distributions are of interest.

To efficiently devise a neural architecture across different data domains, we propose a novel learning task called NASDA (Neural Architecture Search for Domain Adaptation). The ultimate goal of NASDA is to minimize the validation loss of the target domain ($\mathcal{L}_{val}^{t}$). We postulate that a solution to NASDA should not only minimize validation loss of the source domain ($\mathcal{L}_{val}^{s}$), but should also reduce the domain gap between the source and target. To this end, we propose a new NAS learning schema:
\begin{align}
	&\Phi_{\alpha, w} = \mathrm{argmin}_{\alpha} \mathcal{L}^{s}_{val}(w^*(\alpha), \alpha) + \mathrm{disc}(\Phi^*(\mathbf{x}^{s}), \Phi^*(\mathbf{x}^t)) \label{intro_outer} \\
	&\text{s.t.} \quad w^*(\alpha) = \mathrm{argmin}_w \enskip \mathcal{L}_{train}^{s}(w, \alpha) \label{intro_inner}
\end{align}
where $\Phi^*=\Phi_{\alpha, w^*(\alpha)}$, and  $\mathrm{disc}(\Phi^*(\mathbf{x}^{s}), \Phi^*(\mathbf{x}^t))$ denotes the domain discrepancy between the source and target. Note that in unsupervised domain adaptation, $\mathcal{L}_{train}^t$ and $\mathcal{L}_{val}^t$ cannot be computed directly due to the lack of label in the target domain. 

\begin{figure}
    \centering
    \includegraphics[width=\linewidth]{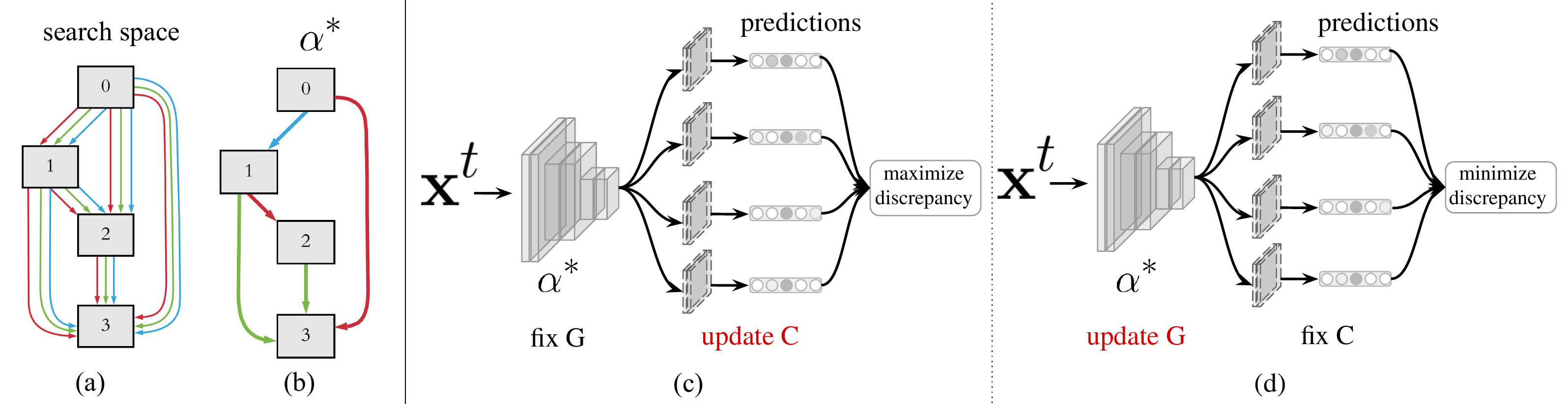}

    \caption{An overview of NASDA: (a) Continuous relaxation of the research space by placing a mixture of the candidate operations on each edge. (b) Inducing the final architecture by joint optimization of the neural architecture parameters $\alpha$ and network weights $w$, supervised by minimizing the validation loss on the source domain and reducing the domain discrepancy. (c)(d) Adversarial training of the derive feature generator $G$ and classifiers $C$.}
    \label{fig_1_overview}
\end{figure}

Inspired by the past works in NAS and unsupervised domain adaptation, we propose in this paper an instantiated NASDA model, which comprises of two training phases, as shown in Figure~\ref{fig_1_overview}. The first is the neural architecture searching phase, aiming to derive an optimal neural architecture ($\alpha^*$), following the learning schema of equation \eqref{intro_outer}\eqref{intro_inner}. Inspired by Differentiable ARchiTecture Search (DARTS)~\cite{darts}, we relax the search space to be continuous so that $\alpha$ can be optimized with respect to $\mathcal{L}^s_{val}$ and $\mathrm{disc}(\Phi(\mathbf{x}^{s}), \Phi(\mathbf{x}^t))$ by gradient descent. Specifically, we enhance the feature transferability by embedding the hidden representations of the task-specific layers to a reproducing kernel Hilbert space where the mean embeddings can be explicitly matched by minimizing $\mathrm{disc}(\Phi(\mathbf{x}^{s}), \Phi(\mathbf{x}^t))$. We use multi-kernel Maximum Mean Discrepancy (MK-MMD)~\cite{gretton2007kernel} to evaluate the domain discrepancy.

The second training phase aims to learn a good feature generator with task-specific loss, based on the derived $\alpha^*$ from the first phase. To establish this goal, we use the derived deep neural network ($\Phi_{\alpha^*}$) as the feature generator ($G$) and devise an adversarial training process between $G$ and a batch of classifiers $C$. The high-level intuition is to first diversify $C$ in the training process, and train $G$ to generate features such that the diversified $C$ can have similar outputs. The training process is similar to Maximum Classifier Discrepancy framework (MCD)~\cite{MCD_2018} except that we extend the dual-classifier in MCD to an ensembling of multiple classifiers. Experiments on standard UDA benchmarks demonstrate the effectiveness of our derived NASDA model in achieving significant improvements over state-of-the-art methods. 

Our contributions of this paper are highlighted as follows:

\begin{itemize}
    \item We formulate a novel dual-objective task of Neural Architecture Search for Domain Adaptation (NASDA), which optimize neural architecture for unsupervised domain adaptation, concerning both source performance objective and transfer learning objective.
    \item We propose an instantiated NASDA model that comprises two training stages, aiming to derive optimal architecture parameters $\alpha^*$ and feature extractor $G$, respectively. We are the first to show the effectiveness of MK-MMD in NAS process specified for domain adaptation.
    \item Extensive experiments on multiple cross-domain recognition tasks demonstrate that NASDA achieves significant improvements over traditional unsupervised domain adaptation models as well as state-of-the-art NAS-based methods.
\end{itemize}

\section{Related Work}
Deep convolutional neural network has been dominating image recognition task. In recent years, many handcrafted architectures have been proposed, including VGG~\cite{vgg}, ResNet~\cite{resnet}, Inception~\cite{szegedy2015going}, etc., all of which verifies the importance of human expertise in network design. Our work bridges domain adaptation and the emerging field of neural architecture search (NAS), a process of automating architecture engineering technique.

\noindent \textbf{Neural Architecture Search} Neural Architecture Search has become the mainstream approach to discover efficient and powerful network structures~\cite{zoph2016neural,zoph2018learning,liu2017hierarchical,liu2018progressive,real2019regularized}. The automatically searched architectures have achieved highly competitive performance in tasks such as image classification~\cite{zoph2016neural,zoph2018learning,liu2017hierarchical,liu2018progressive,real2019regularized}, object detection~\cite{zoph2018learning}, and semantic segmentation~\cite{nas_chen2018searching}. Reinforce learning based NAS methods~\cite{zoph2016neural,tan2019mnasnet,EfficientNet} are usually computational intensive, thus hampering its usage with limited computational budget. To accelerate the search procedure, many techniques has been proposed and they mainly follow four directions: (\textbf{1}) estimating the actual performance with \textit{lower fidelities}. Such lower fidelities include shorter training times~\cite{zoph2018learning,zela2018towards} training on a subset of the data \cite{pmlr-v54-klein17a}, on lower-resolution images~\cite{chrabaszcz2017downsampled}, or with less filters per layer and less cells \cite{zoph2018learning,Real2019AgingEF}. (\textbf{2}) estimating the performance based on the \textit{learning curve extrapolation}. Rawal \textit{et al}~\cite{domhan2015speeding} propose to extrapolate initial learning curves and terminate those predicted to perform poorly and several papers ~\cite{swersky2014freeze,klein2016learning,rawal2018nodes,baker2017accelerating} also leverage architectural hyperparameters to predict which partial learning curves are most promising. (\textbf{3}) \textit{initializing} the novel architectures based on other well-trained architectures. Wei \textit{et al}~\cite{wei2016network} introduce network morphisms to modify an architecture without changing the network objects, resulting in methods that only require a few GPU days~\cite{elsken2017simple,cai2018efficient,jin2019auto,cai2018path}. (\textbf{4}) \textit{one-shot} architecture search. One-shot NAS treats all architectures as different subgraphs of a supergraph and shares weights between architectures that have edges of this supergraph in common~\cite{saxena2016convolutional,brock2017smash,pham2018efficient,liu2018darts,bender2019understanding,cai2018proxylessnas,xie2018snas}. ENAS~\cite{ENAS} learns a RNN controller that samples architectures from the search space and trains the one-shot model using the approximate gradients. DARTS~\cite{darts} places a mixture of candidate operations on each edge of the one-shot model and optimizes the weights of the candidate operations with a continuous relaxation of the search space. Inspired by DARTS~\cite{darts}, our model employs differentiable architecture search to derive the optimal feature extractor for unsupervised domain adaptation.

\noindent \textbf{Domain Adaptation} Unsupervised domain adaptation (UDA) aims to transfer the knowledge learned from one or more labeled source domains to an unlabeled target domain. Various methods have been proposed, including discrepancy-based UDA approaches~\cite{JAN, ddc, ghifary2014domain, peng2017synthetic}, adversary-based approaches~\cite{cogan, adda, ufdn}, and reconstruction-based approaches~\cite{yi2017dualgan, CycleGAN2017, hoffman2017cycada, kim2017learning}. These models are typically designed to tackle single source to single target adaptation. Compared with single source adaptation, multi-source domain adaptation (MSDA) assumes that training data are collected from multiple sources. Originating from the theoretical analysis in~\cite{ben2010theory, Mansour_nips2018, crammer2008learning}, MSDA has been applied to many practical applications~\cite{xu2018deep, duan2012exploiting, domainnet}. Specifically, Ben-David \textit{et al}~\citep{ben2010theory} introduce an $\mathcal{H}\Delta\mathcal{H}$-divergence between the weighted combination of source domains and a target domain. These models are developed using the existing hand-crafted network architecture. This property limits the capacity and versatility of domain adaptation as the backbones to extract the features are fixed. In contrast, we tackle the UDA from a different perspective, not yet considered in the UDA literature. We propose a novel dual-objective model of NASDA, which optimize neural architecture for unsupervised domain adaptation. We are the first to show the effectiveness of MK-MMD in NAS process which is designed specifically for domain adaptation.
\section{Neural Architecture Search for Domain Adaptation}

In unsupervised domain adaptation, we are given a source domain $\mathcal{D}_s = \{(\mathbf{x}_i^s,{\bf y}^s_i)\}_{i=1}^{n_s}$ of $n_s$ labeled examples and a target domain ${{\cal D}_t} = \{ {\bf{x}}_j^t\} _{j = 1}^{{n_t}}$ of $n_t$ unlabeled examples. The source domain and target domain are sampled from joint distributions $P(\mathbf{x}^s, \mathbf{y}^s)$ and $Q(\mathbf{x}^t, \mathbf{y}^t)$, respectively. The goal of this paper is to leverage NAS to derive a deep network $G: {\bf{x}} \mapsto {\bf y}$, which is optimal for reducing the shifts in data distributions across domains, such that the target risk ${\epsilon_t}\left( G \right) = {\mathbb{E} _{\left( {{\mathbf{x}^t},{\bf y}^t} \right) \sim Q}}\left[ {G \left( {\mathbf{x}^t} \right) \ne {\bf y}^t} \right]$ is minimized. We will start by introducing some preliminary background in Section~\ref{preliminary}. We then describe how to incorporate the MK-MMD into the neural architecture searching framework in Section~\ref{nasda_searching}. Finally, we introduce the adversarial training between our derived deep network and a batch of classifiers in Section~\ref{nasda_adversarial_training}. An overview of our model can be seen in Algorithm~\ref{alg_nasda}.

\subsection{Preliminary: DARTS}
\label{preliminary}

In this work, we leverage DARTS~\cite{darts} as our baseline framework. Our goal is to search for a robust cell and apply it to a network that is optimal to achieve domain alignment between $\mathcal{D}_s$ and $\mathcal{D}_t$. Following Zoph \textit{et al}~\cite{zoph2018learning}; Liu \textit{et al}~\cite{darts}, we search for a computation cell as the building block of the final architecture. The final convolutional network for domain adaptation can be stacked from the learned cell. A cell is defined as a directed acyclic graph (DAG) of $L$ nodes, $\{x^{i}\}_{i=1}^{N}$, where each node $x^{(i)}$ is a latent representation and each directed edge $\mathrm{e}^{(i, j)}$ is associated with some operation $o^{(i,j)}$ that transforms $x^{(i)}$. DARTS~\cite{darts} assumes that cells contain two input nodes and a single output node. 
To make the search space continuous, DARTS~\cite{darts} relaxes the categorical choice of a particular operation to a softmax over all possible operations and is thus formulated as:
\begin{equation}
  \Bar{o}^{(i,j)}(x) = \sum_{o\in\mathcal{O}_{}}{\frac{\mathrm{exp}(\alpha_o^{(i,j)})}{\sum_{o'\in\mathcal{O}}\mathrm{exp}(\alpha_{o'}^{(i,j)})}o(x)}
\end{equation}
where $\mathcal{O}$ denotes the set of candidate operations and $i<j$ so that {\it skip-connect} can be applied.
An intermediate node can be represented as $x_j = \sum_{i < j}{o^{(i,j)}(x_i)}$. The task of architecture search then
reduces to learning a set of continuous variables $\alpha = \big\{ \alpha^{(i,j)} \big\}$. At the end of search, a discrete architecture can be obtained by replacing each mixed operation $\bar{o}^{(i,j)}$ with the most likely operation, i.e., $o^{*^{(i,j)}} =\mathrm{argmax}_{o \in \mathcal{O}} \enskip \alpha^{(i,j)}_o$ and $\alpha^*=\{o^{*^{(i,j)}}\}$. 

\subsection{Searching Neural Architecture}
\label{nasda_searching}
Denote by $\mathcal{L}_{train}$ and $\mathcal{L}_{val}$ the training loss and validation loss, respectively. Conventional neural architecture search models aim to derive $\Phi_{\alpha, w}$ by solving a bilevel optimization problem~\cite{anandalingam1992hierarchical}: $\Phi_{\alpha, w}= \argmin_{\alpha} \mathcal{L}_{val}(w^*(\alpha), \alpha) $
	$\text{s.t. } w^*(\alpha) = \mathrm{argmin}_w  \mathcal{L}_{train}(w, \alpha) $. While recent work~\cite{zoph2018learning, liu2017hierarchical} have show promising performance on tasks such as image classification and object detection, the existing models assume that the training data and testing data are sampled from the same distributions. Our goal is to jointly learn the architecture $\alpha$ and the weights $w$ within all the mixed operations (e.g. weights of the convolution filters) so that the derived model $\Phi_{w^*,\alpha^*}$ can transfer knowledge from $\mathcal{D}_s$ to $\mathcal{D}_t$ with some simple domain adapation guidence. Initialized by Equation~\eqref{intro_outer}, we leverage multi-kernel Maximum Mean Discrepancy~\cite{gretton2007kernel} to evaluate $\mathrm{disc}(\Phi^*(\mathbf{x}^{s}), \Phi^*(\mathbf{x}^t)$.

\noindent \textbf{MK-MMD} Denote by $\mathcal{H}_k$ be the reproducing kernel Hilbert space (RKHS) endowed with a characteristic kernel $k$. The \emph{mean embedding} of distribution $p$ in $\mathcal{H}_k$ is a unique element $\mu_k(P)$ such that ${{\mathbf{E}}_{{\mathbf{x}}\sim P}}f\left( {\mathbf{x}} \right) = {\left\langle {f\left( {\mathbf{x}} \right),{\mu _k}\left( P \right)} \right\rangle _{{\mathcal{H}_k}}}$ for all $f \in \mathcal{H}_k$. The MK-MMD $d_k\left( {P,Q} \right)$ between probability distributions $P$ and $Q$ is defined as the RKHS distance between the mean embeddings of $P$ and $Q$. The squared formulation of MK-MMD is defined as
\begin{equation}\label{eqn:MMD}
	d_k^2\left( {P,Q} \right) \triangleq \left\| {{{\mathbf{E}}_P}\left[ {\Phi_{\alpha} \left( {{{\mathbf{x}}^s}} \right)} \right] - {{\mathbf{E}}_Q}\left[ {\Phi_{\alpha} \left( {{{\mathbf{x}}^t}} \right)} \right]} \right\|_{{\mathcal{H}_k}}^2.
\end{equation}

% \vspace{-0.3cm}
\begin{algorithm}[h]			
	\caption{Neural Architecture Search for Domain Adaptation} \label{alg_nasda}
		\hspace*{0.02in}{\bf Phase I: Searching Neural Architecture}

		\begin{algorithmic}[1]
            \State Create a mixed operation $o^{(i,j)}$ parametrized by $\alpha^{(i,j)}$ for each edge $(i,j)$
			\While{not converged}
			\State Update architecture $\alpha$ by 
	$\frac{\partial}{\partial \alpha} \mathcal{L}^s_{val}\bigg(w - \xi \frac{\partial}{\partial w} \mathcal{L}^s_{train}(w, \alpha), \alpha\bigg) +\lambda \frac{\partial}{\partial \alpha} \bigg(\hat{d}_k^2\left( \Phi(\mathbf{x}^s),\Phi(\mathbf{x}^t)\right)\bigg)$\;

            \State Update weights $w$ by descending $\frac{\partial}{\partial w}  \mathcal{L}^s_{train}(w, \alpha)$\;
            \EndWhile
            \State Derive the final architecture based on the learned $\alpha^*$.
            
        \end{algorithmic}
        \noindent\noindent\hspace*{0.02in}{\bf Phase II: Adversarial Training for Domain Adaptation}
        \begin{algorithmic}[1]
            \State Stack feature generator $G$ based on $\alpha^*$, initialize classifiers $C$
			\While{not converged}
			\State Step one: Train $G$ and $C$ with $\mathcal{L}^s(\mathbf{x}^{s},\mathbf{y}^{s}) = -{\mathbb{E}_{(\mathbf{x}^{s},\mathbf{y}^{s})\sim\mathcal{D}^s}}\sum_{k=1}^{K}{\mathds{1}_{[k=\mathbf{y}^{s}]}}\log p({\mathbf{y}^{s}}|{\mathbf{x}^{s}})$
% 			\State Step one: Train $G$ and $C$ with $\mathcal{L}^s(\mathbf{x}^{s},\mathbf{y}^{s}) =-{\mathbb{E}_{(\mathbf{x}^{s},\mathbf{y}^{s})\sim\mathcal{D}^s}}\sum_{k=1}^{K}{\ma1_{[k=\mathbf{y}^{s}]}}$
            \State Step two: Fix G, train C with loss: $\mathcal{L}^s(\mathbf{x}^{s},\mathbf{y}^{s})- \mathcal{L}_{\rm adv}(\mathbf{x}^{t}) (Eq.~\ref{eq:sensitivity})$
            \State Step three: Fix C, train G with loss: $\mathcal{L}_{\rm adv}(\mathbf{x}^{t})$
            \EndWhile
        \end{algorithmic}

\end{algorithm}   

In this paper, we consider the case of combining Gaussian kernels with injective functions $f_\Phi$, where 
${k}(x, x') = \exp(-\|f_\Phi(x)-f_\Phi(x)'\|^2)$. 
Inspired by Long \textit{et al}~\cite{long2015}, the characteristic kernel associated with the feature map $\Phi$, $k\left( {{{\mathbf{x}}^s},{{\mathbf{x}}^t}} \right) = \left\langle {\Phi \left( {{{\mathbf{x}}^s}} \right),\Phi \left( {{{\mathbf{x}}^t}} \right)} \right\rangle $, is defined as the convex combination of $n$ positive semidefinite kernels $\{k_u\}$,
\begin{equation}\label{eqn:MK}
	\mathcal{K} \triangleq \left\{ {k = \sum\limits_{u = 1}^n {{\beta _u}{k_u}} :\sum\limits_{u = 1}^n {{\beta _u}}  = 1,{\beta _u} \geqslant 0}, {\forall u} \right\},
\end{equation}
where the constraints on $\{\beta_u\}$ are imposed to guarantee that the $k$ is characteristic. In practice we use finite samples from distributions to estimate MMD distance. 
Given $\mathbf{X}^s = \{\mathbf{x}^s_1, \cdots, \mathbf{x}^s_m\} \sim P$ and $\mathbf{X}^t = \{\mathbf{x}^t_1, \cdots, \mathbf{x}^t_m\} \sim Q$, one estimator of $d_k^2(P, Q)$ is
\begin{equation}\label{eqn_practical}
    \hat{d}_k^2(P, Q) = \frac{1}{{m \choose 2}}\sum_{i\neq i'} k(\mathbf{x^s}_i, \mathbf{x^s}_i') - \frac{2}{{m \choose 2}}\sum_{i\neq j} k(\mathbf{x}_i^s,
	\mathbf{x}_j^t) + \frac{1}{{m \choose 2}}\sum_{j\neq j'} k(\mathbf{x^t}_j, \mathbf{x^t}_j').
\end{equation}

The merit of multi-kernel MMD lies in its differentiability such that it can be easily incorporated into the deep network. However, the computation of the $\hat{d}_k^2(P, Q)$ incurs a complexity of $O(m^2)$, which is undesirable in the differentiable architecture search framework. In this paper, we use the unbiased estimation of MK-MMD~\cite{gretton2012kernel} which can be computed with linear complexity.  

\noindent \textbf{NAS for Domain Adaptation} 
Denote by $\mathcal{L}_{train}^s$ and $\mathcal{L}_{val}^s$ the training loss and validation loss on the source domain, respectively. Both losses are affected by the architecture $\alpha$ as well as by the weights $w$ in the network. The goal for NASDA is to find $\alpha^{*}$ that minimizes the validation loss $\mathcal{L}^t_{val}(w^{*}, \alpha^{*})$ on the target domain, where the weights $w^{*}$ associated with the architecture are obtained by minimizing the training loss $w^* = \mathrm{argmin}_w\ \mathcal{L}^s_{train}(w, \alpha^*)$. Due to the lack of labels in the target domain, it is prohibitive to compute $\mathcal{L}^t_{val}$ directly, hampering the assumption of previous gradient-based NAS algorithms~\cite{darts,chen2019progressive}. Instead, we derive $\alpha^*$ by minimizing the validation loss $\mathcal{L}^s_{val}(w^{*}, \alpha^{*})$ on the source domain plus the domain discrepancy, $\mathrm{disc}(\Phi(\mathbf{x}^s),\Phi(\mathbf{x}^t))$, as shown in equation \eqref{intro_outer}.

Inspired by the gradient-based hyperparameter optimization~\cite{franceschi2018bilevel,pedregosa2016hyperparameter,maclaurin2015gradient}, we set the architecture parameters $\alpha$ as a special type of hyperparameter. This implies a bilevel optimization problem~\cite{anandalingam1992hierarchical} with $\alpha$ as the upper-level variable and $w$ as the lower-level variable. In practice, we utilize the MK-MMD to evaluate the domain discrepancy. The optimization can be summarized as follows:
\begin{align}
	&\Phi_{\alpha, w} = \mathrm{argmin}_{\alpha} \bigg(\mathcal{L}^s_{val}(w^*(\alpha), \alpha)+ \lambda \hat{d}_k^2\left( {\Phi(\mathbf{x}^s),\Phi(\mathbf{x}^t)} \right) \bigg) \label{eq:outer} \\
	&\text{s.t.  }  w^*(\alpha) = \mathrm{argmin}_w \enskip \mathcal{L}^s_{train}(w, \alpha) \label{eq:inner}
\end{align}
where $\lambda$ is the trade-off hyperparameter between the source validation loss and the MK-MMD loss.

\noindent \textbf{Approximate Architecture Search} Equation~\eqref{eq:outer}\eqref{eq:inner} imply that directly optimizing the architecture gradient is prohibitive due to the expensive inner optimization. Inspired by DARTS~\cite{darts}, we approximate $w^*(\alpha)$ by adapting $w$ using only a single training step, without solving the optimization in Equation~\eqref{eq:inner} by training until convergence. This idea has been adopted and proven to be effective in meta-learning for model transfer~\cite{maml}, gradient-based hyperparameter tuning~\cite{luketina2016scalable} and unrolled generative adversarial networks. We therefore propose a simple approximation scheme as follows:
\begin{align}
&\frac{\partial}{\partial \alpha} \bigg(\mathcal{L}^s_{val}(w^*(\alpha), \alpha)+ \lambda \hat{d}_k^2\left( {\Phi(\mathbf{x}^s),\Phi(\mathbf{x}^t)} \right)\bigg) \nonumber\\
\approx &\frac{\partial}{\partial \alpha} \mathcal{L}^s_{val}\bigg(w - \xi \frac{\partial}{\partial w} \mathcal{L}^s_{train}(w, \alpha), \alpha\bigg) +\lambda \frac{\partial}{\partial \alpha} \bigg(\hat{d}_k^2\left( \Phi(\mathbf{x}^s),\Phi(\mathbf{x}^t)\right)\bigg) \label{eq:single}
\end{align}
where $w - \xi \frac{\partial}{\partial w} \mathcal{L}^s_{train}(w, \alpha)$ denotes weight for one-step forward model and $\xi$ is the learning rate for a step of inner optimization. Note Equation \eqref{eq:single} reduces to $\nabla_\alpha \mathcal{L}_{val}(w, \alpha)$ if $w$ is already a local optimum for the inner optimization and thus $\nabla_w \mathcal{L}_{train}(w, \alpha) = 0$.

The second term of Equation~\eqref{eq:single} can be computed directly with some forward and backward passes. For the first term, applying chain rule to the approximate architecture gradient yields
\begin{align}
\scriptsize{
	\frac{\partial}{\partial \alpha} \mathcal{L}^s_{val}(w', \alpha) - \xi \bigg( \frac{\partial^2}{\partial\alpha\partial w} \mathcal{L}^s_{train}(w, \alpha) \frac{\partial}{\partial w'} \mathcal{L}^s_{val}(w', \alpha) \bigg) 
	}
        \label{eq:arch-grad}
\end{align}
where $w' = w - \xi \frac{\partial}{\partial w} \mathcal{L}_{train}(w, \alpha)$. The expression above contains an expensive matrix-vector product 
in its second term. We leverage the central difference approximation to reduce the computation complexity. Specifically,  
let $\eta$ be a small scalar and
$w^\pm = w \pm \eta \frac{\partial}{\partial w'} \mathcal{L}^s_{val}(w', \alpha)$. Then:
\begin{equation}
\frac{\partial^2}{\partial\alpha\partial w} \mathcal{L}^s_{train}(w, \alpha) \frac{\partial}{\partial w'} \mathcal{L}^s_{val}(w', \alpha) 
	\approx
	\frac{\frac{\partial}{\partial \alpha} \mathcal{L}_{train}(w^+, \alpha) - \frac{\partial}{\partial \alpha} \mathcal{L}_{train}(w^-, \alpha)}{2 \eta}
\end{equation}

Evaluating the central difference only requires two forward passes for the weights and two backward passes for $\alpha$, reducing the complexity from quadratic to linear.

\vspace{-0.2cm}
\subsection{Adversarial Training for Domain Adaptation}
\label{nasda_adversarial_training}
\vspace{-0.2cm}
% \todo{the w in section 3.2 and 3.3 is a bit confusing now and should}

By neural architecture searching from Section~\ref{nasda_searching}, we have derived the optimal cell structure ($\alpha^*$) for domain adaptation. We then stack the cells to derive our feature generator $G$. In this section, we describe how do we consolidate $G$ by an adversarial training of $G$ and the classifiers $C$. Assume $C$ includes $N$ independent classifiers {$\{C^{(i)}\}_{i=1}^N$} and denote $p_i(\mathbf{y}|\mathbf{x})$ as the $K$-way propabilistic outputs of $C^{(i)}$, where $K$ is the category number.

The high-level intuition is to consolidate the feature generator $G$ such that it can make the diversified $C$ generate similar outputs. To this end, our training process include three steps: (1) train $G$ and $C$ on $\mathcal{D}_s$ to obtain task-specific features, (2) fix $G$ and train $C$ to make {$\{C^{(i)}\}_{i=1}^N$} have diversified output, (3) fix $C$ and train $G$ to minimize the output discrepancy between $C$. Related techniques have been used in Saito et al~\cite{MCD_2018}; Kumar et al~\cite{NIPS2018_8146}.

\newcommand{\mymin}{\mathop{\rm min}\limits}
\newcommand{\mymax}{\mathop{\rm max}\limits}
\newcommand{\1}{\mbox{1}\hspace{-0.25em}\mbox{l}}

First, we train both $G$ and $C$ to classify the source samples correctly with cross-entropy loss. This step is crucial as it enables $G$ and $C$ to extract the task-specific features. The training objective is $\mymin_{G,C} \mathcal{L}^s(\mathbf{x}^{s},\mathbf{y}^{s})$ and the loss function is defined as follows:
  \begin{equation}
   \mathcal{L}^s(\mathbf{x}^{s},\mathbf{y}^{s}) = -{\mathbb{E}_{(\mathbf{x}^{s},\mathbf{y}^{s})\sim\mathcal{D}^s}}\sum_{k=1}^{K}{\1_{[k=\mathbf{y}^{s}]}}\log p({\mathbf{y}^{s}}|{\mathbf{x}^{s}})
   \label{eq:crossentropy}
  \end{equation}

In the second step, we are aiming to diversify $C$. To establish this goal, we fix $G$ and train $C$ to increase the discrepancy of $C$'s output. To avoid mode collapse (\textit{e.g.} $C^{(1)}$ outputs all zeros and $C^{(2)}$ output all ones), we add $\mathcal{L}^s(\mathbf{x}^{s},\mathbf{y}^{s})$ as a regularizer in the training process. The high-level intuition is that we do not expect $C$ to forget the information learned in the first step in the training process. The training objective is $\mymin_{C} \mathcal{L}^s(\mathbf{x}^{s},\mathbf{y}^{s})-\mathcal{L}_{\rm adv}(\mathbf{x}^{t})$, where the adversarial loss is defined as:

\begin{equation}
  \mathcal{L}_{\rm adv}(\mathbf{x}^{t}) = {\mathbb{E}_{{\mathbf{x}^{t}}\sim \mathcal{D}^{t}}}  \sum_{i=1}^{N-1}\sum_{j=i+1}^{N}\|(p_i(\mathbf{y}|\mathbf{x}^t)- p_j(\mathbf{y}|\mathbf{x}^t)\|_1
  \label{eq:sensitivity}
\end{equation}

In the last step, we are trying to consolidate the feature generator $G$ by training $G$ to extract generalizable representations such that the discrepancy of $C$'s output is minimized. To achieve this goal, we fix the diversified classifiers $C$ and train $G$ with the adversarial loss (defined in Equation~\ref{eq:sensitivity}). The training objective is  
 $\mymin_{G} \mathcal{L}_{\rm adv}(\mathbf{x}_{t})$. In the testing phase, the final prediction is the average of the $N$ classifiers.

\section{Experiments}
\label{sec:exp}
\begin{figure*}[t!]
 \vspace{-0.4cm}
    \begin{minipage}{\hsize}
    \centering
    \subfigure[\small Normal cells (left) and Reduce cells (right) for STL$\rightarrow$CIFAR10]{
    \includegraphics[width=0.45\linewidth]{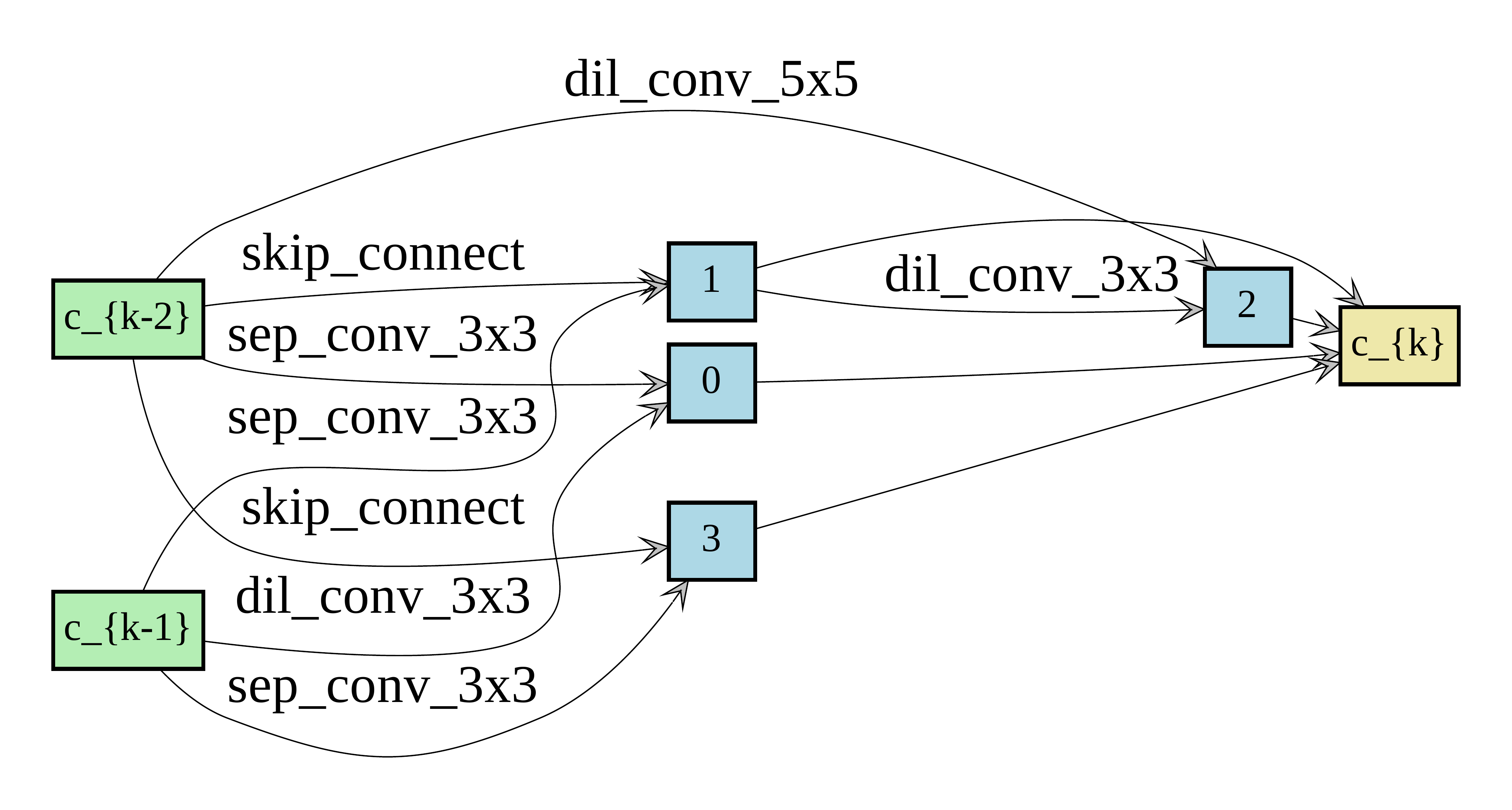}
    \includegraphics[width=0.53\linewidth]{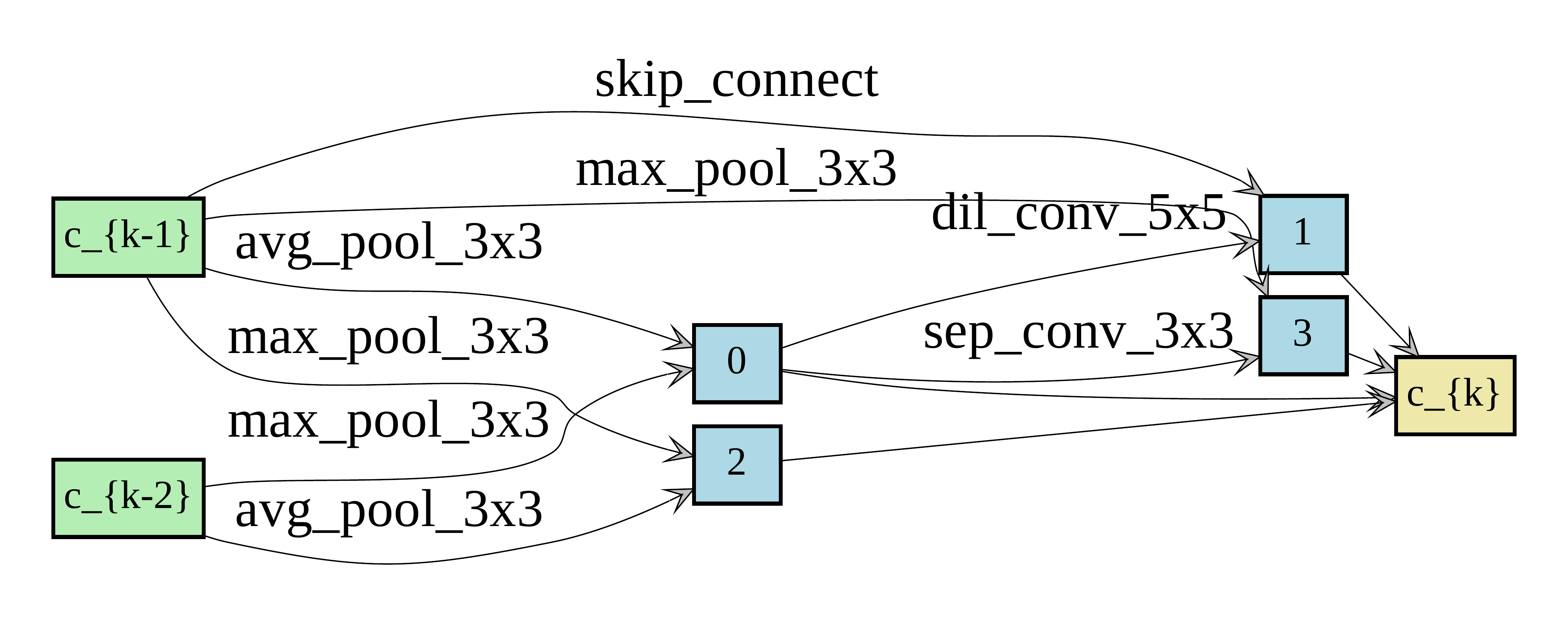}}
    
    \subfigure[\small Normal cells (upper) and Reduce cells (lower) for MNIST$\rightarrow$USPS]{
    \includegraphics[width=0.65\linewidth]{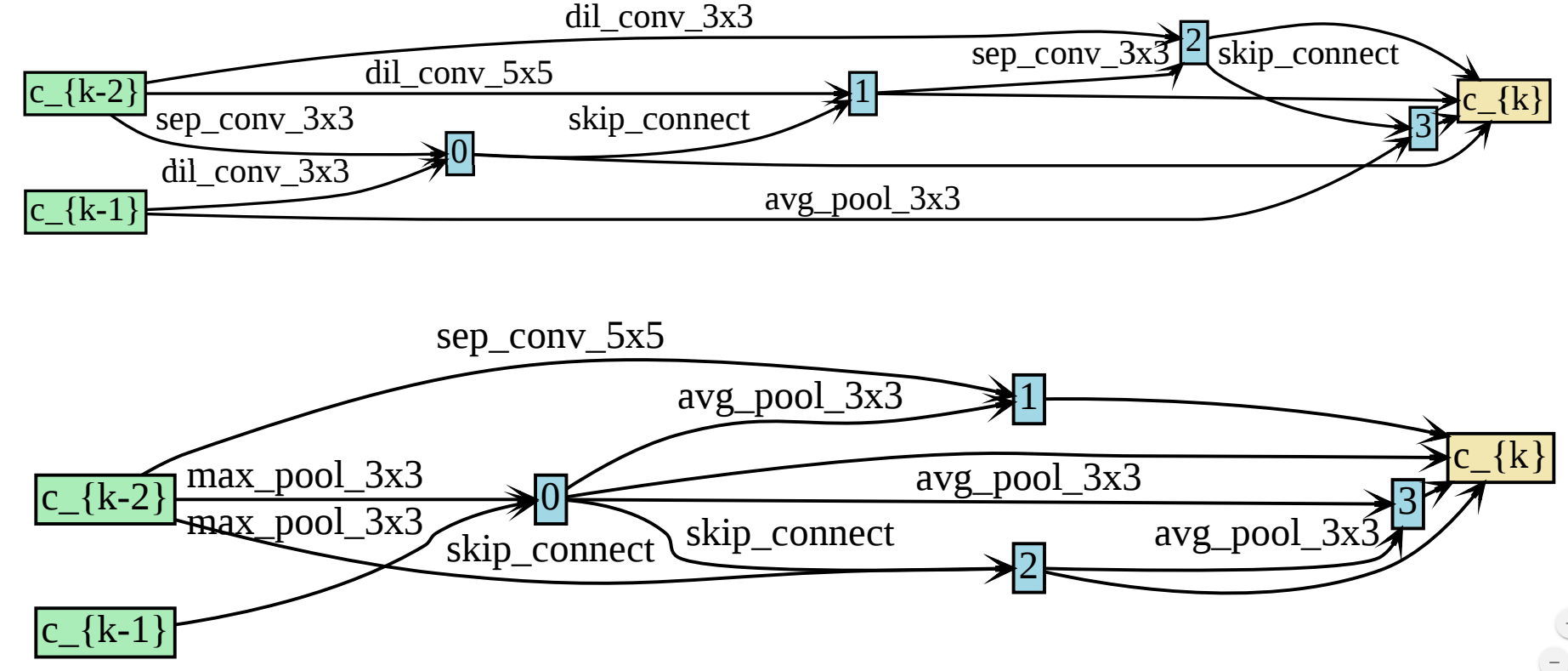}
    }
    \begin{minipage}{0.28\linewidth}%
    \vspace{-30mm}
    \scriptsize
    % \subfloat{
    \centering
    \begin{tabular}{c p{0.5cm} *{8}{p{1.2cm}}}
    % \begin{tabular}{cccc}
  	\toprule
    Method & Params (M) & Search Cost (GPU days)\\
    \midrule
    NASNet~\cite{zoph2018learning} & 3.3 & 1,800 \\
    AmoebaNet~\cite{shah2018amoebanet} & 3.2  & 3,150  \\
    PNAS~\cite{liu2018progressive} & 3.2 & 225 \\
    DARTS~\cite{darts}  &  3.3 & 1.5 \\
    SNAS~\cite{xie2018snas} & 2.8 & 1.5 \\
    PDARTS\cite{chen2019progressive}  &  3.4 & 0.3 \\
    \midrule
    \textbf{NASDA}  &  2.7 & 0.3 \\
    \bottomrule
  \end{tabular}
    % }
    \caption
    *{\small (c) Network architecture statistics comparison}
    \label{fig_c}
    \end{minipage}
    \caption{(a) Neural architecture for STL$\rightarrow$CIFAR10 task. (b) Neural architecture for MNIST$\rightarrow$USPS results. (c) Comparison between our NASDA model and state-of-the-art NAS models.}
    \label{fig:arch}
    \end{minipage}
    \vspace{-0.6cm}
\end{figure*}

We compare the proposed NASDA model with many state-of-the-art UDA baselines on multiple benchmarks. In the main paper, we only report major results; more details are provided in the supplementary material. All of our experiments are implemented in the PyTorch platform.

In the architecture search phase, we use $\lambda$=1 for all the searching experiments. We leverage the ReLU-Conv-BN order for convolutional operations, and each separable convolution is always applied twice \citep{zoph2018learning, real2019regularized, liu2018progressive}. Our search space $\mathcal{O}$ includes the following operations: $3\times3$ and $5\times5$ separable convolutions, $3\times3$ and $5\times5$ dilated separable convolutions, $3\times3$ max pooling, identity, and $zero$. Our convolutional cell consists of $N=7$ nodes. Cells located at the $\frac{1}{3}$ and $\frac{2}{3}$ of the total depth of the network are reduction cells. The architecture encoding therefore is $(\alpha_{normal}, \alpha_{reduce})$, where $\alpha_{normal}$ is shared by all the normal cells and $\alpha_{reduce}$ is shared by all the reduction cells. In the adversarial training phase, we set $N$=4 for all the experiments. 

\vspace{-0.2cm}
\subsection{Setup}
\textbf{Digits} We investigate three digits datasets: \textbf{MNIST}, \textbf{USPS}, and Street View House Numbers (\textbf{SVHN}). We adopt the evaluation protocol of CyCADA \cite{hoffman2017cycada} with three transfer tasks: USPS to MNIST (\textbf{U $\rightarrow$ M}), MNIST to USPS (\textbf{M $\rightarrow$ U}), and SVHN to MNIST (\textbf{S $\rightarrow$ M}). We train our model using the training sets: MNIST (60,000), USPS (7,291), standard SVHN train (73,257).

\textbf{STL$\rightarrow$CIFAR10} Both CIFAR10~\cite{cifar10} and STL~\cite{stl10} are both 10-class image datasets. These two datasets contain nine overlapping classes. We remove the `frog' class in CIFAR10 and the `monkey' class in STL datasets as they have no equivalent in the other dataset, resulting in a 9-class problem. The STL images were down-scaled to 32$\times$32 resolution to match that of CIFAR10.

\textbf{SYN SIGNS$\rightarrow$GTSRB} We evaluated the adaptation from synthetic traffic signs dataset (SYN SIGNS dataset~\cite{synthetic_sign}) to real-world signs dataset (GTSRB dataset~\cite{GTSRB}). These datasets contain 43 classes. 

We compare our NASDA model with state-of-the-art domain adaptation methods: Deep Adaptation Network (\textbf{DAN})~\cite{long2015}, Domain
Adversarial Neural Network (\textbf{DANN})~\cite{DANN}, Domain Separation Network (\textbf{DSN})~\cite{bousmalis2016domain}, Coupled Generative Adversarial Networks (\textbf{CoGAN})~\cite{cogan}, Maximum Classifier Discrepancy (\textbf{MCD})~\cite{MCD_2018}, Generate to Adapt (\textbf{G2A})~\cite{g2a}, Stochastic Neighborhood Embedding (\textbf{d-SNE})~\cite{xu2019d}, Associative Domain Adaptation (\textbf{ASSOC})~\cite{ASSC}.

\vspace{-0.2cm}
\subsection{Empirical Results}
\vspace{-0.2cm}
\noindent \textbf{Neural Architecture Search Results} We show the neural architecture search results in Figure~\ref{fig:arch}. We also show that our NASDA model contains less parameters and takes less time to converge compared with state-of-the-art NAS architectures. One interesting finding is that our NASDA contains more sequential connections in both Normal and Reduce cells when trained on MNIST$\rightarrow$USPS.
%\afterpage{
\begin{table}[t!]
  \vspace{-6pt}
  \addtolength{\tabcolsep}{6pt}
  \centering 
  \caption{Accuracy (\%) on {Digits} and {Traffic Signs} for unsupervised domain adaptation.}
  \label{tab_exp}
%   \vspace{-5pt}
  \resizebox{\textwidth}{!}{%
  \begin{tabular}{ccccc|cc}
  	\toprule
    Method & M $\rightarrow$ U & U $\rightarrow$ M & S $\rightarrow$ M & Avg(digits) & Method & SYN SIGNS $\rightarrow$ GTSRB \\
    % GTRSB  $\rightarrow$ SYNSig & 
    % CIFAR $\rightarrow$ STL  &
    \midrule % non NAS method
    DAN~\cite{li2017mmd}  & 81.1 & - & 71.1 & 76.1 &  DAN~\cite{li2017mmd}&91.1\\
    DANN~\cite{DANN} & 85.1 & 73.0 & 71.1 & 76.4 & DANN~\cite{DANN}  & 88.7  \\
    DSN~\cite{dsn} & $91.3\footnote[1]{used partial dataset}$ & - & 82.7 & - & DSN~\cite{dsn}  & 93.1 \\
    % ADDA~\cite{adda} & $89.4^*$  & 90.1 & 76.0 & - & \\
    CoGAN~\cite{cogan} & $91.2^*$ & 89.1 & - & - & CORAL~\cite{sun2016deep} & 86.9 \\
    
    MCD~\cite{MCD_2018}  & 94.2 & 94.1 & 96.2 & 94.8 & MCD~\cite{MCD_2018} & 94.4 \\ %https://arxiv.org/pdf/1909.05288v1.pdf
    G2A~\cite{g2a} & 95.0 & 90.8 & 92.4 &  92.7&  ASSOC~\cite{ASSC} & 82.8\\
    SBADA-GAN~\cite{sbada} & 95.3 & 97.6 & 76.1 & 89.7 & SRDA-RAN~\cite{srda} & 93.6\\
    d-SNE~\cite{xu2019d} & \textbf{99.0} & 98.7 & 96.5 &  \underline{98.1} & DADRL~\cite{DADRL} & 94.6\\
    
    \midrule % NAS method
    % NASNet & 98.4 & 99.0 & 98.8 & \\
    % AmoebaNet &  & 98.9 &  &  & \\
    % DARTS  & 98.1 & 98.8 & 98.8 &  &\\
    % PDARTS  & 98.2 & 98.4  & 98.7 &  &  \\
    \textbf{NASDA}  &  98.0 & \textbf{98.7}  & \textbf{98.6} & \textbf{98.4} &\textbf{NASDA} & \textbf{96.7} \\
    \bottomrule
  \end{tabular}
  }
%   \vspace{-13pt}
\end{table}

    %  &91.1\\
    %   & 88.7 \\
    %  & 93.1   \\
    %  & 86.9 \\
    %  & 93.6 \\
    % & 94.2   \\
    % ASSOC~\cite{ASSC} & 82.8 \\
    % DADRL~\cite{DADRL} & 94.6\\
\begin{figure*}[t!]
\vspace{-0.4cm}
    \begin{minipage}{\hsize}
    \centering
    
    \subfigure[\scriptsize T-SNE embedding for 4 classifiers' weights ]{
    \label{fig_c_4}
    \includegraphics[width=0.3\linewidth]{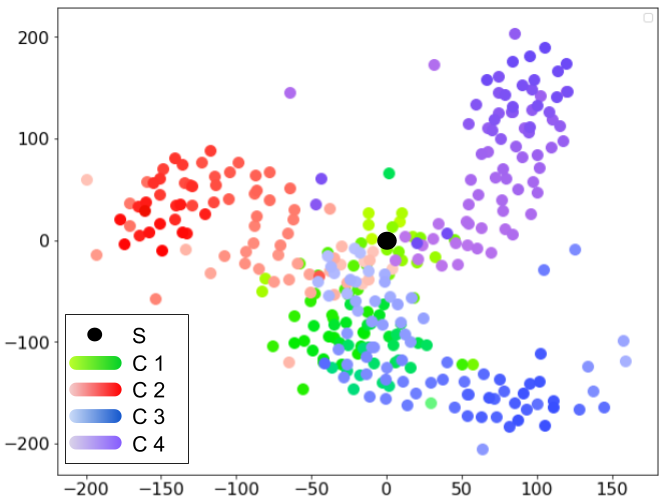}
    \hspace{5pt}
    }
    \subfigure[\scriptsize T-SNE embedding for 5 classifiers' weights ]{
    \includegraphics[width=0.3\linewidth]{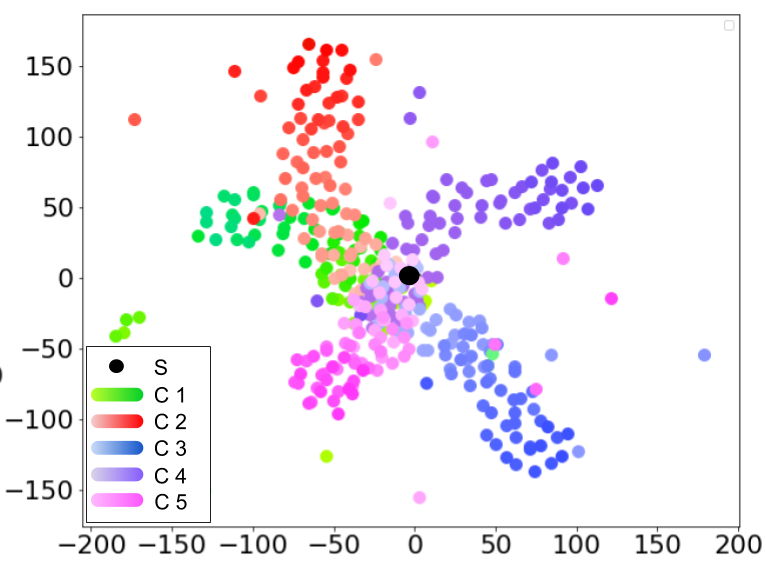}
    \hspace{5pt}
    \label{fig_c_5}
    }
    \subfigure[\scriptsize Accuracy \textit{v.s.} Classifier Number ]{
    \includegraphics[width=0.3\linewidth]{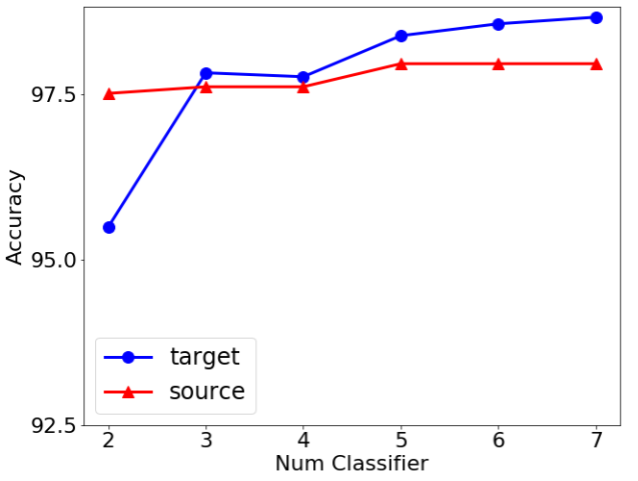}
    \label{fig_c_acciracy_vs_c}
    }
    \vspace{-0.2cm}
    \caption{\footnotesize {(\textbf{a})(\textbf{b})} We show the classifiers are diverged from each other in USPS$\rightarrow$MNIST task. (\textbf{c}) The relation between the source and target accuracy \textit{v.s.} the number of classifiers in USPS$\rightarrow$MNIST task.}
    \end{minipage}
    \vspace{-0.5cm}
\end{figure*}
%}

\noindent \textbf{Unsupervised Domain Adaptation Results} The UDA results for Digits and SYN SIGNS$\rightarrow$GTSRB are reported in Table~\ref{tab_exp}, with results of baselines directly reported from the original papers if the protocol is the same (numbers with $^*$ indicates training on partial data). The NASDA model achieves a \textbf{98.4\%} average accuracy for Digits dataset, outperforming other baselines. For SYN SIGNS$\rightarrow$GTSRB task, our model gets comparable results with state-of-the-art baselines. The results demonstrate the effectiveness of our NASDA model on small images.

The UDA results on the STL$\rightarrow$CIFAR10 recognition task are reported in Table~\ref{tab_traffic}. Our model achieves a performance of \textbf{76.8\%}, outperforming all the baselines. To compare our search neural architecture with previous NAS models, we replace the neural architecture we used in $G$ with other NAS models.

\begin{wraptable}{r}{0.4\textwidth}
\scriptsize{
  \addtolength{\tabcolsep}{3pt}
  
  \centering 
  \caption[width=0.2\textwidth]{\footnotesize{Accuracy} (\%) \scriptsize{on {STL $\rightarrow$ CIFAR10}.}}
  \label{tab_traffic}
  \resizebox{0.4\textwidth}{!}{%
  \begin{tabular}{cc}
  	\toprule
     Method & STL $\rightarrow$ CIFAR10\\
    \midrule % non NAS method
    DANN~\cite{DANN}  & 56.9  \\
    MCD~\cite{MCD_2018} & 69.2  \\
    DWT~\cite{roy2019unsupervised} & 71.2 \\
    SE~\cite{french2017self} & 74.2 \\ 
    G2A~\cite{g2a} & 72.8\\
    VADA~\cite{shu2018dirt} & 73.5\\
    DIRT-T~\cite{shu2018dirt} & \underline{75.3}\\
    \midrule % NAS method
    NASNet+\textit{Phase II}~\cite{zoph2018learning} & 67.3 \\
    AmoebaNet+\textit{Phase II}~\cite{shah2018amoebanet} & 67.0 \\
    DARTS+\textit{Phase II}~\cite{darts}  & 68.8 \\
PDARTS+\textit{Phase II}~\cite{chen2019progressive}  &  66.0 \\
    \midrule
    \textbf{NASDA} & \textbf{76.8} \\
    \bottomrule
  \end{tabular}
  }
  \vspace{-0.5cm}
}
\end{wraptable}
Other training settings in the second phase are identical to our model. As such, we derive \textbf{NASNet}+\textit{Phase II}~\cite{zoph2018learning}, \textbf{AmoebaNet}+\textit{Phase II}~\cite{shah2018amoebanet} , \textbf{DARTS}+\textit{Phase II}~\cite{darts}, and \textbf{PDARTS}+\textit{Phase II}~\cite{chen2019progressive} models. The results in Table~\ref{tab_traffic} demonstrate that our model outperform other NAS based model by a large margin, demonstrating the effectiveness of our model in unsupervised domain adaptation.

\textbf{Analysis} To dive deeper into the training process of our NASDA model, we plot in Figure~\ref{fig_c_4}-\ref{fig_c_5} the T-SNE embedding of the weights of $C$ in USPS$\rightarrow$MNIST. This is achieved by recording the weights of all the classifiers for each epoch. The black dot indicates epoch zero, which is the common starting point. The color from light to dart corresponds to the epoch number from small to large. The T-SNE plots clearly show that the classifiers are diverged from each other, demonstrating the effectiveness of the second step of our NASDA training described in Section~\ref{nasda_searching}. 

To explore the effect of the number of classifiers ($N$) on the final results, we plot the source and target accuracy \textit{v.s.} the number of classifiers for USPS$\rightarrow$MNIST in Figure~\ref{fig_c_acciracy_vs_c}. The plot shows that with more classifiers, the accuracy of target domain increases significantly, while the accuracy of the source domain improves by a small margin. This is an interesting finding since it shows that the diversified classifiers boost the performance on the target domain. However, since the computation complexity of the adversarial loss defined by Equation~\eqref{eq:sensitivity} is $O(N^2)$, the computation cost will increase quadratically when we increase $N$. We select $N$=4 for all our experiments, a trade-off between the performance and the computation efficiency.

\vspace{-0.3cm}
\section{Conclusion}
\vspace{-0.3cm}
In this paper, we first formulate a novel dual-objective task of Neural Architecture Search for Domain Adaptation (NASDA) to invigorate the design of transfer-aware network architectures.
Towards tackling the NASDA task, we have proposed a novel learning framework that leverages MK-MMD to guide the neural architecture search process. Instead of aligning the features from existing handcrafted backbones, our model directly searches for the optimal neural architecture specific for domain adaptation. Furthermore, we have introduced the ways to consolidate the feature generator, which is stacked from the searched architecture, in order to boost the UDA performance. Extensive empirical evaluations on UDA benchmarks have demonstrated the efficacy of the proposed model against several state-of-the-art domain adaptation algorithms.

\clearpage
\section{Broader Impacts}

 \textbf{Data side} The efficacy of deep learning algorithms highly relies on abundant labeled training data. However, annotating the large-scale dataset is tedious. In addition, labeling costs can be prohibitively expensive for some applications, requiring specialized expertise (labeling X-rays for medical diagnosis) or significant manual effort (pixel-level annotations for semantic segmentation). As such, reusing the learned knowledge from cheaper source of labels (e.g. existing labeled datasets, synthetic data) to generalize to new tasks and datasets is an critical but unsolved challenge.

\textbf{Model side} On the other hand, discovering state-of-the-art neural network architectures requires substantial effort of human experts. Recently, there has been a growing interest in developing algorithmic solutions to automate the manual process of architecture design. However, these models assume that the training and testing domain are sampled from the same distribution, neglecting the scenario where two data domains or multiple distributions are of interest. 

Our work focuses on cost-efficient generalization by identifying a small subset of target data that will, once labeled, lead to good target performance. Specifically, we anticipate our line of work to automatically searching for the good deep backbone for knowledge transfer task. In terms of impact on society, this could mean that computer vision systems are able to better handle novel deployments and are less susceptible to dataset bias. For example, our system could adapt a traffic sign recognition system which is trained on simulated data to the real traffic sign recognition task. Although we do not experiment on fairness applications, domain adaptation has also been shown to improve the fairness of face recognition systems across race/gender.

 Other negative impacts of our research on society are harder to predict, but it suffers from the same issues as most deep learning algorithms. These include adversarial attacks, privacy concerns and lack of interpretability, as well as other negative effects of increased automation. For example, it is not easy to interpret why the searched architecture is better to perform specific domain adaptation tasks.

\bibliographystyle{unsrt}
\bibliography{refs}

\begin{thebibliography}{10}

\bibitem{datashift_book2009}
Joaquin Quionero-Candela, Masashi Sugiyama, Anton Schwaighofer, and Neil~D.
  Lawrence.
\newblock {\em Dataset Shift in Machine Learning}.
\newblock The MIT Press, 2009.

\bibitem{JAN}
Mingsheng Long, Han Zhu, Jianmin Wang, and Michael~I. Jordan.
\newblock Deep transfer learning with joint adaptation networks.
\newblock In {\em Proceedings of the 34th International Conference on Machine
  Learning, {ICML} 2017, Sydney, NSW, Australia, 6-11 August 2017}, pages
  2208--2217, 2017.

\bibitem{ddc}
Eric Tzeng, Judy Hoffman, Ning Zhang, Kate Saenko, and Trevor Darrell.
\newblock Deep domain confusion: Maximizing for domain invariance.
\newblock {\em arXiv preprint arXiv:1412.3474}, 2014.

\bibitem{ghifary2014domain}
Muhammad Ghifary, W~Bastiaan Kleijn, and Mengjie Zhang.
\newblock Domain adaptive neural networks for object recognition.
\newblock In {\em Pacific Rim international conference on artificial
  intelligence}, pages 898--904. Springer, 2014.

\bibitem{peng2017synthetic}
Xingchao Peng and Kate Saenko.
\newblock Synthetic to real adaptation with generative correlation alignment
  networks.
\newblock In {\em 2018 {IEEE} Winter Conference on Applications of Computer
  Vision, {WACV} 2018, Lake Tahoe, NV, USA, March 12-15, 2018}, pages
  1982--1991, 2018.

\bibitem{long2015}
Mingsheng Long, Yue Cao, Jianmin Wang, and Michael Jordan.
\newblock Learning transferable features with deep adaptation networks.
\newblock In Francis Bach and David Blei, editors, {\em Proceedings of the 32nd
  International Conference on Machine Learning}, volume~37 of {\em Proceedings
  of Machine Learning Research}, pages 97--105, Lille, France, 07--09 Jul 2015.
  PMLR.

\bibitem{SunS16a}
Baochen Sun and Kate Saenko.
\newblock Deep {CORAL:} correlation alignment for deep domain adaptation.
\newblock {\em CoRR}, abs/1607.01719, 2016.

\bibitem{xu2019adversarial}
Minghao Xu, Jian Zhang, Bingbing Ni, Teng Li, Chengjie Wang, Qi~Tian, and
  Wenjun Zhang.
\newblock Adversarial domain adaptation with domain mixup.
\newblock {\em arXiv preprint arXiv:1912.01805}, 2019.

\bibitem{cogan}
Ming-Yu Liu and Oncel Tuzel.
\newblock Coupled generative adversarial networks.
\newblock In {\em Advances in neural information processing systems}, pages
  469--477, 2016.

\bibitem{adda}
Eric Tzeng, Judy Hoffman, Kate Saenko, and Trevor Darrell.
\newblock Adversarial discriminative domain adaptation.
\newblock In {\em Computer Vision and Pattern Recognition (CVPR)}, volume~1,
  page~4, 2017.

\bibitem{ufdn}
Alexander~H. Liu, Yen{-}Cheng Liu, Yu{-}Ying Yeh, and Yu{-}Chiang~Frank Wang.
\newblock A unified feature disentangler for multi-domain image translation and
  manipulation.
\newblock {\em CoRR}, abs/1809.01361, 2018.

\bibitem{DANN}
Yaroslav Ganin and Victor Lempitsky.
\newblock Unsupervised domain adaptation by backpropagation.
\newblock In Francis Bach and David Blei, editors, {\em Proceedings of the 32nd
  International Conference on Machine Learning}, volume~37 of {\em Proceedings
  of Machine Learning Research}, pages 1180--1189, Lille, France, 07--09 Jul
  2015. PMLR.

\bibitem{MCD_2018}
Kuniaki Saito, Kohei Watanabe, Yoshitaka Ushiku, and Tatsuya Harada.
\newblock Maximum classifier discrepancy for unsupervised domain adaptation.
\newblock In {\em The IEEE Conference on Computer Vision and Pattern
  Recognition (CVPR)}, June 2018.

\bibitem{long2018NIPS_CDAN}
Mingsheng Long, Zhangjie Cao, Jianmin Wang, and Michael~I Jordan.
\newblock Conditional adversarial domain adaptation.
\newblock In {\em Advances in Neural Information Processing Systems}, pages
  1640--1650, 2018.

\bibitem{yi2017dualgan}
Zili Yi, Hao~(Richard) Zhang, Ping Tan, and Minglun Gong.
\newblock Dualgan: Unsupervised dual learning for image-to-image translation.
\newblock In {\em ICCV}, pages 2868--2876, 2017.

\bibitem{CycleGAN2017}
Jun-Yan Zhu, Taesung Park, Phillip Isola, and Alexei~A Efros.
\newblock Unpaired image-to-image translation using cycle-consistent
  adversarial networks.
\newblock In {\em Computer Vision (ICCV), 2017 IEEE International Conference
  on}, 2017.

\bibitem{hoffman2017cycada}
Judy Hoffman, Eric Tzeng, Taesung Park, Jun-Yan Zhu, Phillip Isola, Kate
  Saenko, Alexei Efros, and Trevor Darrell.
\newblock {C}y{CADA}: Cycle-consistent adversarial domain adaptation.
\newblock In Jennifer Dy and Andreas Krause, editors, {\em Proceedings of the
  35th International Conference on Machine Learning}, volume~80 of {\em
  Proceedings of Machine Learning Research}, pages 1989--1998,
  Stockholmsmässan, Stockholm Sweden, 10--15 Jul 2018. PMLR.

\bibitem{kim2017learning}
Taeksoo Kim, Moonsu Cha, Hyunsoo Kim, Jung~Kwon Lee, and Jiwon Kim.
\newblock Learning to discover cross-domain relations with generative
  adversarial networks.
\newblock In Doina Precup and Yee~Whye Teh, editors, {\em Proceedings of the
  34th International Conference on Machine Learning}, volume~70 of {\em
  Proceedings of Machine Learning Research}, pages 1857--1865, International
  Convention Centre, Sydney, Australia, 06--11 Aug 2017. PMLR.

\bibitem{zoph2016neural}
Barret Zoph and Quoc~V Le.
\newblock Neural architecture search with reinforcement learning.
\newblock {\em ICLR}, 2017.

\bibitem{Real2019AgingEF}
Esteban Real, Alok Aggarwal, Yanping Huang, and Quoc~V. Le.
\newblock Aging evolution for image classifier architecture search.
\newblock In {\em AAAI 2019}, 2019.

\bibitem{anandalingam1992hierarchical}
G~Anandalingam and Terry~L Friesz.
\newblock Hierarchical optimization: An introduction.
\newblock {\em Annals of Operations Research}, 34(1):1--11, 1992.

\bibitem{zoph2018learning}
Barret Zoph, Vijay Vasudevan, Jonathon Shlens, and Quoc~V Le.
\newblock Learning transferable architectures for scalable image recognition.
\newblock In {\em Proceedings of the IEEE conference on computer vision and
  pattern recognition}, pages 8697--8710, 2018.

\bibitem{liu2017hierarchical}
Hanxiao Liu, Karen Simonyan, Oriol Vinyals, Chrisantha Fernando, and Koray
  Kavukcuoglu.
\newblock Hierarchical representations for efficient architecture search.
\newblock {\em ICLR}, 2018.

\bibitem{liu2018progressive}
Chenxi Liu, Barret Zoph, Maxim Neumann, Jonathon Shlens, Wei Hua, Li-Jia Li,
  Li~Fei-Fei, Alan Yuille, Jonathan Huang, and Kevin Murphy.
\newblock Progressive neural architecture search.
\newblock In {\em Proceedings of the European Conference on Computer Vision
  (ECCV)}, pages 19--34, 2018.

\bibitem{real2019regularized}
Esteban Real, Alok Aggarwal, Yanping Huang, and Quoc~V Le.
\newblock Regularized evolution for image classifier architecture search.
\newblock In {\em Proceedings of the aaai conference on artificial
  intelligence}, volume~33, pages 4780--4789, 2019.

\bibitem{darts}
Hanxiao Liu, Karen Simonyan, and Yiming Yang.
\newblock Darts: Differentiable architecture search.
\newblock {\em International Conference on Learning Representations}, 2019.

\bibitem{gretton2007kernel}
Arthur Gretton, Karsten~M Borgwardt, Malte Rasch, Bernhard Sch{\"o}lkopf, and
  Alex~J Smola.
\newblock A kernel method for the two-sample-problem.
\newblock In {\em Advances in neural information processing systems}, pages
  513--520, 2007.

\bibitem{vgg}
Karen Simonyan and Andrew Zisserman.
\newblock Very deep convolutional networks for large-scale image recognition.
\newblock {\em CoRR}, abs/1409.1556, 2014.

\bibitem{resnet}
Kaiming He, Xiangyu Zhang, Shaoqing Ren, and Jian Sun.
\newblock Deep residual learning for image recognition.
\newblock In {\em Proceedings of the IEEE conference on computer vision and
  pattern recognition}, pages 770--778, 2016.

\bibitem{szegedy2015going}
Christian Szegedy, Wei Liu, Yangqing Jia, Pierre Sermanet, Scott Reed, Dragomir
  Anguelov, Dumitru Erhan, Vincent Vanhoucke, and Andrew Rabinovich.
\newblock Going deeper with convolutions.
\newblock In {\em Proceedings of the IEEE conference on computer vision and
  pattern recognition}, pages 1--9, 2015.

\bibitem{nas_chen2018searching}
Liang-Chieh Chen, Maxwell Collins, Yukun Zhu, George Papandreou, Barret Zoph,
  Florian Schroff, Hartwig Adam, and Jon Shlens.
\newblock Searching for efficient multi-scale architectures for dense image
  prediction.
\newblock In {\em Advances in neural information processing systems}, pages
  8699--8710, 2018.

\bibitem{tan2019mnasnet}
Mingxing Tan, Bo~Chen, Ruoming Pang, Vijay Vasudevan, Mark Sandler, Andrew
  Howard, and Quoc~V Le.
\newblock Mnasnet: Platform-aware neural architecture search for mobile.
\newblock In {\em Proceedings of the IEEE Conference on Computer Vision and
  Pattern Recognition}, pages 2820--2828, 2019.

\bibitem{EfficientNet}
Mingxing Tan and Quoc Le.
\newblock {E}fficient{N}et: Rethinking model scaling for convolutional neural
  networks.
\newblock In Kamalika Chaudhuri and Ruslan Salakhutdinov, editors, {\em
  Proceedings of the 36th International Conference on Machine Learning},
  volume~97 of {\em Proceedings of Machine Learning Research}, pages
  6105--6114, Long Beach, California, USA, 09--15 Jun 2019. PMLR.

\bibitem{zela2018towards}
Arber Zela, Aaron Klein, Stefan Falkner, and Frank Hutter.
\newblock Towards automated deep learning: Efficient joint neural architecture
  and hyperparameter search.
\newblock {\em ICML 2018 Workshop on AutoML (AutoML 2018)}, 2018.

\bibitem{pmlr-v54-klein17a}
Aaron Klein, Stefan Falkner, Simon Bartels, Philipp Hennig, and Frank Hutter.
\newblock {Fast Bayesian Optimization of Machine Learning Hyperparameters on
  Large Datasets}.
\newblock In Aarti Singh and Jerry Zhu, editors, {\em Proceedings of the 20th
  International Conference on Artificial Intelligence and Statistics},
  volume~54 of {\em Proceedings of Machine Learning Research}, pages 528--536,
  Fort Lauderdale, FL, USA, 20--22 Apr 2017. PMLR.

\bibitem{chrabaszcz2017downsampled}
Patryk Chrabaszcz, Ilya Loshchilov, and Frank Hutter.
\newblock A downsampled variant of imagenet as an alternative to the cifar
  datasets.
\newblock {\em arXiv preprint arXiv:1707.08819}, 2017.

\bibitem{domhan2015speeding}
Tobias Domhan, Jost~Tobias Springenberg, and Frank Hutter.
\newblock Speeding up automatic hyperparameter optimization of deep neural
  networks by extrapolation of learning curves.
\newblock In {\em Twenty-Fourth International Joint Conference on Artificial
  Intelligence}, 2015.

\bibitem{swersky2014freeze}
Kevin Swersky, Jasper Snoek, and Ryan~Prescott Adams.
\newblock Freeze-thaw bayesian optimization.
\newblock {\em arXiv preprint arXiv:1406.3896}, 2014.

\bibitem{klein2016learning}
Aaron Klein, Stefan Falkner, Jost~Tobias Springenberg, and Frank Hutter.
\newblock Learning curve prediction with bayesian neural networks.
\newblock {\em ICLR}, 2017.

\bibitem{rawal2018nodes}
Aditya Rawal and Risto Miikkulainen.
\newblock From nodes to networks: Evolving recurrent neural networks.
\newblock {\em arXiv preprint arXiv:1803.04439}, 2018.

\bibitem{baker2017accelerating}
Bowen Baker, Otkrist Gupta, Ramesh Raskar, and Nikhil Naik.
\newblock Accelerating neural architecture search using performance prediction.
\newblock {\em NIPS Workshop on Meta-Learning}, 2017.

\bibitem{wei2016network}
Tao Wei, Changhu Wang, Yong Rui, and Chang~Wen Chen.
\newblock Network morphism.
\newblock In {\em International Conference on Machine Learning}, pages
  564--572, 2016.

\bibitem{elsken2017simple}
Thomas Elsken, Jan-Hendrik Metzen, and Frank Hutter.
\newblock Simple and efficient architecture search for convolutional neural
  networks.
\newblock {\em NIPS Workshop on Meta-Learning}, 2017.

\bibitem{cai2018efficient}
Han Cai, Tianyao Chen, Weinan Zhang, Yong Yu, and Jun Wang.
\newblock Efficient architecture search by network transformation.
\newblock In {\em Thirty-Second AAAI conference on artificial intelligence},
  2018.

\bibitem{jin2019auto}
Haifeng Jin, Qingquan Song, and Xia Hu.
\newblock Auto-keras: An efficient neural architecture search system.
\newblock In {\em Proceedings of the 25th ACM SIGKDD International Conference
  on Knowledge Discovery \& Data Mining}, pages 1946--1956, 2019.

\bibitem{cai2018path}
Han Cai, Jiacheng Yang, Weinan Zhang, Song Han, and Yong Yu.
\newblock Path-level network transformation for efficient architecture search.
\newblock {\em International Conference on Machine Learning}, 2018.

\bibitem{saxena2016convolutional}
Shreyas Saxena and Jakob Verbeek.
\newblock Convolutional neural fabrics.
\newblock In {\em Advances in Neural Information Processing Systems}, pages
  4053--4061, 2016.

\bibitem{brock2017smash}
Andrew Brock, Theodore Lim, James~M Ritchie, and Nick Weston.
\newblock Smash: one-shot model architecture search through hypernetworks.
\newblock {\em NIPS Workshop on Meta-Learning}, 2017.

\bibitem{pham2018efficient}
Hieu Pham, Melody~Y Guan, Barret Zoph, Quoc~V Le, and Jeff Dean.
\newblock Efficient neural architecture search via parameter sharing.
\newblock {\em International Conference on Machine Learning}, 2018.

\bibitem{liu2018darts}
Hanxiao Liu, Karen Simonyan, and Yiming Yang.
\newblock Darts: Differentiable architecture search.
\newblock {\em International Conference on Learning Representations}, 2019.

\bibitem{bender2019understanding}
Gabriel Bender.
\newblock Understanding and simplifying one-shot architecture search.
\newblock {\em International Conference on Machine Learning}, 2018.

\bibitem{cai2018proxylessnas}
Han Cai, Ligeng Zhu, and Song Han.
\newblock Proxylessnas: Direct neural architecture search on target task and
  hardware.
\newblock {\em International Conference on Learning Representations}, 2019.

\bibitem{xie2018snas}
Sirui Xie, Hehui Zheng, Chunxiao Liu, and Liang Lin.
\newblock Snas: stochastic neural architecture search.
\newblock {\em International Conference on Learning Representations}, 2019.

\bibitem{ENAS}
Hieu Pham, Melody~Y Guan, Barret Zoph, Quoc~V Le, and Jeff Dean.
\newblock Efficient neural architecture search via parameter sharing.
\newblock {\em International Conference on Machine Learning}, 2018.

\bibitem{ben2010theory}
Shai Ben-David, John Blitzer, Koby Crammer, Alex Kulesza, Fernando Pereira, and
  Jennifer~Wortman Vaughan.
\newblock A theory of learning from different domains.
\newblock {\em Machine learning}, 79(1-2):151--175, 2010.

\bibitem{Mansour_nips2018}
Yishay Mansour, Mehryar Mohri, Afshin Rostamizadeh, and A~R.
\newblock Domain adaptation with multiple sources.
\newblock In D.~Koller, D.~Schuurmans, Y.~Bengio, and L.~Bottou, editors, {\em
  Advances in Neural Information Processing Systems 21}, pages 1041--1048.
  Curran Associates, Inc., 2009.

\bibitem{crammer2008learning}
Koby Crammer, Michael Kearns, and Jennifer Wortman.
\newblock Learning from multiple sources.
\newblock {\em Journal of Machine Learning Research}, 9(Aug):1757--1774, 2008.

\bibitem{xu2018deep}
Ruijia Xu, Ziliang Chen, Wangmeng Zuo, Junjie Yan, and Liang Lin.
\newblock Deep cocktail network: Multi-source unsupervised domain adaptation
  with category shift.
\newblock In {\em Proceedings of the IEEE Conference on Computer Vision and
  Pattern Recognition}, pages 3964--3973, 2018.

\bibitem{duan2012exploiting}
Lixin Duan, Dong Xu, and Shih-Fu Chang.
\newblock Exploiting web images for event recognition in consumer videos: A
  multiple source domain adaptation approach.
\newblock In {\em Computer Vision and Pattern Recognition (CVPR), 2012 IEEE
  Conference on}, pages 1338--1345. IEEE, 2012.

\bibitem{domainnet}
Xingchao Peng, Qinxun Bai, Xide Xia, Zijun Huang, Kate Saenko, and Bo~Wang.
\newblock Moment matching for multi-source domain adaptation.
\newblock In {\em Proceedings of the IEEE International Conference on Computer
  Vision}, pages 1406--1415, 2019.

\bibitem{gretton2012kernel}
Arthur Gretton, Karsten~M Borgwardt, Malte~J Rasch, Bernhard Sch{\"o}lkopf, and
  Alexander Smola.
\newblock A kernel two-sample test.
\newblock {\em Journal of Machine Learning Research}, 13(Mar):723--773, 2012.

\bibitem{chen2019progressive}
Xin Chen, Lingxi Xie, Jun Wu, and Qi~Tian.
\newblock Progressive differentiable architecture search: Bridging the depth
  gap between search and evaluation.
\newblock In {\em Proceedings of the IEEE International Conference on Computer
  Vision}, pages 1294--1303, 2019.

\bibitem{franceschi2018bilevel}
Luca Franceschi, Paolo Frasconi, Saverio Salzo, Riccardo Grazzi, and
  Massimilano Pontil.
\newblock Bilevel programming for hyperparameter optimization and
  meta-learning.
\newblock {\em ICML}, 2018.

\bibitem{pedregosa2016hyperparameter}
Fabian Pedregosa.
\newblock Hyperparameter optimization with approximate gradient.
\newblock 2016.

\bibitem{maclaurin2015gradient}
Dougal Maclaurin, David Duvenaud, and Ryan Adams.
\newblock Gradient-based hyperparameter optimization through reversible
  learning.
\newblock In {\em International Conference on Machine Learning}, pages
  2113--2122, 2015.

\bibitem{maml}
Chelsea Finn, Pieter Abbeel, and Sergey Levine.
\newblock Model-agnostic meta-learning for fast adaptation of deep networks.
\newblock In Doina Precup and Yee~Whye Teh, editors, {\em Proceedings of the
  34th International Conference on Machine Learning}, volume~70 of {\em
  Proceedings of Machine Learning Research}, pages 1126--1135, International
  Convention Centre, Sydney, Australia, 06--11 Aug 2017. PMLR.

\bibitem{luketina2016scalable}
Jelena Luketina, Mathias Berglund, Klaus Greff, and Tapani Raiko.
\newblock Scalable gradient-based tuning of continuous regularization
  hyperparameters.
\newblock In {\em International conference on machine learning}, pages
  2952--2960, 2016.

\bibitem{NIPS2018_8146}
Abhishek Kumar, Prasanna Sattigeri, Kahini Wadhawan, Leonid Karlinsky, Rogerio
  Feris, Bill Freeman, and Gregory Wornell.
\newblock Co-regularized alignment for unsupervised domain adaptation.
\newblock In S.~Bengio, H.~Wallach, H.~Larochelle, K.~Grauman, N.~Cesa-Bianchi,
  and R.~Garnett, editors, {\em Advances in Neural Information Processing
  Systems 31}, pages 9345--9356. Curran Associates, Inc., 2018.

\bibitem{shah2018amoebanet}
Syed Asif~Raza Shah, Wenji Wu, Qiming Lu, Liang Zhang, Sajith Sasidharan, Phil
  DeMar, Chin Guok, John Macauley, Eric Pouyoul, Jin Kim, et~al.
\newblock Amoebanet: An sdn-enabled network service for big data science.
\newblock {\em Journal of Network and Computer Applications}, 119:70--82, 2018.

\bibitem{cifar10}
Alex Krizhevsky, Geoffrey Hinton, et~al.
\newblock Learning multiple layers of features from tiny images.
\newblock 2009.

\bibitem{stl10}
Adam Coates, Andrew Ng, and Honglak Lee.
\newblock An analysis of single-layer networks in unsupervised feature
  learning.
\newblock In {\em Proceedings of the fourteenth international conference on
  artificial intelligence and statistics}, pages 215--223, 2011.

\bibitem{synthetic_sign}
Boris Moiseev, Artem Konev, Alexander Chigorin, and Anton Konushin.
\newblock Evaluation of traffic sign recognition methods trained on
  synthetically generated data.
\newblock In Jacques Blanc-Talon, Andrzej Kasinski, Wilfried Philips, Dan
  Popescu, and Paul Scheunders, editors, {\em Advanced Concepts for Intelligent
  Vision Systems}, pages 576--583, Cham, 2013. Springer International
  Publishing.

\bibitem{GTSRB}
Johannes Stallkamp, Marc Schlipsing, Jan Salmen, and Christian Igel.
\newblock The {G}erman {T}raffic {S}ign {R}ecognition {B}enchmark: A
  multi-class classification competition.
\newblock In {\em IEEE International Joint Conference on Neural Networks},
  pages 1453--1460, 2011.

\bibitem{bousmalis2016domain}
Konstantinos Bousmalis, George Trigeorgis, Nathan Silberman, Dilip Krishnan,
  and Dumitru Erhan.
\newblock Domain separation networks.
\newblock In {\em Advances in neural information processing systems}, pages
  343--351, 2016.

\bibitem{g2a}
Swami Sankaranarayanan, Yogesh Balaji, Carlos~D Castillo, and Rama Chellappa.
\newblock Generate to adapt: Aligning domains using generative adversarial
  networks.
\newblock In {\em Proceedings of the IEEE Conference on Computer Vision and
  Pattern Recognition}, pages 8503--8512, 2018.

\bibitem{xu2019d}
Xiang Xu, Xiong Zhou, Ragav Venkatesan, Gurumurthy Swaminathan, and Orchid
  Majumder.
\newblock d-sne: Domain adaptation using stochastic neighborhood embedding.
\newblock In {\em Proceedings of the IEEE Conference on Computer Vision and
  Pattern Recognition}, pages 2497--2506, 2019.

\bibitem{ASSC}
P.~{Haeusser}, T.~{Frerix}, A.~{Mordvintsev}, and D.~{Cremers}.
\newblock Associative domain adaptation.
\newblock In {\em 2017 IEEE International Conference on Computer Vision
  (ICCV)}, pages 2784--2792, 2017.

\bibitem{li2017mmd}
Chun-Liang Li, Wei-Cheng Chang, Yu~Cheng, Yiming Yang, and Barnab{\'a}s
  P{\'o}czos.
\newblock Mmd gan: Towards deeper understanding of moment matching network.
\newblock In {\em Advances in Neural Information Processing Systems}, pages
  2203--2213, 2017.

\bibitem{dsn}
Konstantinos Bousmalis, George Trigeorgis, Nathan Silberman, Dilip Krishnan,
  and Dumitru Erhan.
\newblock Domain separation networks.
\newblock In {\em Advances in Neural Information Processing Systems}, pages
  343--351, 2016.

\bibitem{sun2016deep}
Baochen Sun and Kate Saenko.
\newblock Deep coral: Correlation alignment for deep domain adaptation.
\newblock In {\em European conference on computer vision}, pages 443--450.
  Springer, 2016.

\bibitem{sbada}
Paolo Russo, Fabio~M Carlucci, Tatiana Tommasi, and Barbara Caputo.
\newblock From source to target and back: symmetric bi-directional adaptive
  gan.
\newblock In {\em Proceedings of the IEEE Conference on Computer Vision and
  Pattern Recognition}, pages 8099--8108, 2018.

\bibitem{srda}
Guanyu Cai, Yuqin Wang, and Lianghua He.
\newblock Learning smooth representation for unsupervised domain adaptation,
  2019.

\bibitem{DADRL}
V.~{Tran} and C.~{Huang}.
\newblock Domain adaptation meets disentangled representation learning and
  style transfer.
\newblock In {\em 2019 IEEE International Conference on Systems, Man and
  Cybernetics (SMC)}, pages 2998--3005, 2019.

\bibitem{roy2019unsupervised}
Subhankar Roy, Aliaksandr Siarohin, Enver Sangineto, Samuel~Rota Bulo, Nicu
  Sebe, and Elisa Ricci.
\newblock Unsupervised domain adaptation using feature-whitening and consensus
  loss.
\newblock In {\em Proceedings of the IEEE Conference on Computer Vision and
  Pattern Recognition}, pages 9471--9480, 2019.

\bibitem{french2017self}
Geoffrey French, Michal Mackiewicz, and Mark Fisher.
\newblock Self-ensembling for visual domain adaptation.
\newblock {\em arXiv preprint arXiv:1706.05208}, 2017.

\bibitem{shu2018dirt}
Rui Shu, Hung~H Bui, Hirokazu Narui, and Stefano Ermon.
\newblock A dirt-t approach to unsupervised domain adaptation.
\newblock {\em arXiv preprint arXiv:1802.08735}, 2018.

\end{thebibliography}

% \clearpage
\newpage
\section{Supplementary Material}
\renewcommand{\thefootnote}{\arabic{footnote}}
\subsection{Notations}
We conclude all the notations we used in the paper in Table~\ref{tab:notations}. Specifically, the notations with superscription ``s'' and ``t'' indicate they are from source domain and target domain, respectively.

\begin{table}[h!]
    \centering
    \begin{tabular}{|c|c|}
    \hline
        Notation & Description  \\
         $\alpha$& Neural architecture parameters\\
         $w$ & Neural network weights \\
         $\Phi$ & Machine learning model\\
         $P / Q$ & Source / Target distribution \\
                   $G$ & Feature generator in phase II \\
          $C$ & Classifier in phase II \\
         $\mathcal{L}_{train}$ / $\mathcal{L}_{val}$  & Training / Validation loss \\
         $\mathbf{x}$ ($\mathbf{x}^s$, $\mathbf{x}^t$)  & Training data (source, target) \\
         $\mathbf{y}$ ($\mathbf{y}^s$, $\mathbf{y}^t$)  & Labels (source, target)\\
          $\mathcal{D}_s$ / $\mathcal{D}_t$ & Source / Target Domain \\
          $o^{(i,j)}$ / $e^{(i,j)}$ & Operation / Edge from node i to j in NAS Graph\\
          $x^{(i)}$ & $i$-th node in a neural cell\\
          $L$ & Number of nodes in a neural cell \\
          $d_k^2$ / $\hat{d}_k^2$ & MK-MMD / Empirical MK-MMD \\
          $k$ / $\mathcal{K}$ & Kernel / Kernels \\
          $m$ & Number of data used to compute MK-MMD\\
          $\xi$ & Learning rate of inner optimization\\
          $\lambda$ & Trade-off parameters\\
          $K$ & Class number\\
          $p(y|x)$ & Probabilistic outputs of classifiers\\
          $N$ & Number of Classifiers \\
          
         \hline
    \end{tabular}
    \caption{We conclude all the notations we used in the paper. }
    \label{tab:notations}
\end{table}

\subsection{Data Statistics}
\label{app:datasets}

The datasets used in this paper are described in Table~\ref{tab:datasets}. Specifically, the \textbf{USPS}, \textbf{MNIST}, \textbf{SVHN}, \textbf{CIFARO-10}, \textbf{STL} datasets are from \textsc{torchvision}\footnote{\url{https://pytorch.org/docs/stable/torchvision/datasets.html}}. The \textbf{SYN SIGNS}\footnote{\url{http://graphics.cs.msu.ru/en/node/1337}} and GTSRB~\footnote{\url{http://benchmark.ini.rub.de/?section=gtsrb}} dataset are downloaded from their official websites.  

\begin{table}[h!t]
\begin{center}
\footnotesize
\begin{tabular}{lllllll}
\hline
{} &              \# train &     \# test &      \# classes &   Target &     Resolution &      Channels \\
\hline

USPS &   7,291 &        2,007 &        10 &           Digits &     $16\times16$ &    Mono     \\    
MNIST &           60,000 &       10,000 &       10 &           Digits &     $28\times28$ &    Mono     \\
SVHN &            73,257 &       26,032 &       10 &           Digits &     $32\times32$ &    RGB     \\
CIFAR-10 &        50,000 &       10,000 &       10 &           Object ID &    $32\times32$ &    RGB     \\
STL &    5,000 &        8,000 &        10 &           Object ID &    $96\times96$ &    RGB     \\
SYN SIGNS &       100,000 &      -- &           43 &           Traffic signs &  $40\times40$ &    RGB     \\
GTSRB &           32,209 &       12,630 &       43 &           Traffic signs &  \emph{varies} &   RGB     \\

\hline
\end{tabular}

\caption{Statistics of datasets we used in our paper.}
\label{tab:datasets}
\end{center}
\end{table}

\textbf{Data Preparation} Some of the experiments that involved datasets described in Table~\ref{tab:datasets} required additional data preparation in order to match the resolution and format of the input samples and match the classification target. These additional steps will now be described.

\textbf{STL $\rightarrow$ CIFAR-10} CIFAR-10 and STL are both 10-class image datasets. The STL images were down-scaled to $32\times32$ resolution to match that of CIFAR-10. The `frog' class in CIFAR-10 and the `monkey' class in STL were removed as they have no equivalent in the other dataset, resulting in a 9-class problem with 10\% less samples in each dataset.

\textbf{Syn-Signs $\rightarrow$ GTSRB} GTSRB is composed of images that vary in size and come with annotations that provide region of interest (bounding box around the sign) and ground truth classification. We extracted the region of interest from each image and scaled them to a resolution of $40\times40$ to match those of Syn-Signs.

\textbf{SVHN $\rightarrow$ MNIST} The MNIST images were padded to $32\times32$ resolution and converted to RGB by replicating the greyscale channel into the three RGB channels to match the format of SVHN.

\subsection{Additional Experiment Results}
In the paper, we report the result of SYN SIGNS$\rightarrow$GTSRB as 93.7\%. After some grid search for the training hyper-parameters, we found that our model can actually achieve an accuracy of \textbf{96.7\%}, outperforming state-of-the-art baselines. All the results in Table~\ref{tab_exp_sup} can be reproduced by the code we attached in the supplementary material.

\begin{table}[h!]
  \vspace{-6pt}
  \addtolength{\tabcolsep}{6pt}
  \centering 
  \caption{Accuracy (\%) on {Digits} and {Traffic Signs} for unsupervised domain adaptation.}
  \label{tab_exp_sup}
  \resizebox{\textwidth}{!}{%
  \begin{tabular}{ccccc|cc}
  	\toprule
    Method & M $\rightarrow$ U & U $\rightarrow$ M & S $\rightarrow$ M & Avg(digits) & Method & SYN SIGNS $\rightarrow$ GTSRB \\
    \midrule 
    DAN~\cite{li2017mmd}  & 81.1 & - & 71.1 & 76.1 &  DAN~\cite{li2017mmd}&91.1\\
    DANN~\cite{DANN} & 85.1 & 73.0 & 71.1 & 76.4 & DANN~\cite{DANN}  & 88.7  \\
    DSN~\cite{dsn} & $91.3\footnote[1]{used partial dataset}$ & - & 82.7 & - & DSN~\cite{dsn}  & 93.1 \\
    CoGAN~\cite{cogan} & $91.2^*$ & 89.1 & - & - & CORAL~\cite{sun2016deep} & 86.9 \\
    
    MCD~\cite{MCD_2018}  & 94.2 & 94.1 & 96.2 & 94.8 & MCD~\cite{MCD_2018} & 94.4 \\
    G2A~\cite{g2a} & 95.0 & 90.8 & 92.4 &  92.7&  ASSOC~\cite{ASSC} & 82.8\\
    SBADA-GAN~\cite{sbada} & 95.3 & 97.6 & 76.1 & 89.7 & SRDA-RAN~\cite{srda} & 93.6\\
    d-SNE~\cite{xu2019d} & \textbf{99.0} & 98.7 & 96.5 &  \underline{98.1} & DADRL~\cite{DADRL} & \textbf{94.6}\\
    
    \midrule 
    \textbf{NASDA}  &  98.0 & \textbf{98.7}  & \textbf{98.6} & \textbf{98.4} &\textbf{NASDA} & \textbf{96.7} \\
    \bottomrule
  \end{tabular}
  }
\end{table}

\subsection{ML reproducibility}

We have submitted the code for all the experiments in the supplementary material. We will briefly describe the details about the experiments. 

\textbf{Hyper-parameter} We set $\lambda$ to be 1 for all the experiments. The range of learning rate we considered is between 2e-4 to 0.25. We adopt grid search to select the best hyper-parameters. All the hyper-parameters used to generate results can be viewed in the code. 

\textbf{Measure} For all the quantitative results in the paper, we use accuracy as the measurement.

\textbf{Average runtime} For Phase I in our model, \textit{i.e.} searching the neural architecture, our model takes 0.3 GPU day to find the optimal architecture. For Phase II, our model typically takes about two days to converge.

\textbf{Computing infrastructure} Our code is based on Pytorch 1.2.0 and Torchvision 0.4.2. All other descriptions can be found in the readme file in the code. 

\end{document}